\def\year{2022}\relax
\pdfoutput=1
\documentclass[letterpaper]{article} 
\usepackage{aaai20}  
\usepackage{times}  
\usepackage{helvet} 
\usepackage{courier}  
\usepackage[hyphens]{url}  
\usepackage{graphicx} 
\usepackage{graphicx}  
\usepackage{amsmath}
\usepackage{amsthm}
\usepackage{amssymb}
\title{Deep Semi-supervised Learning with Double-Contrast of Features and Semantics}
\author{Quan Feng$^{1}$, Jiayu Yao$^{1}$, Zhison Pan$^{2}$*, Guojun Zhou$^{3}$\\
~\\
\textit{\small $^1$ Nanjing University of Aeronautics and Astronautics, China}\\
\textit{\small $^2$Army Engineering University, China}\\
\textit{\small $^3$Yulin Normal University, China }\\
~\\
    E-mail:  $\{$ fengquan, jiayu\_yao, panzs\}@nuaa.edu.cn, ylsyzgj@126.com\\
}

\begin{document}

\maketitle

\begin{abstract}
In recent years, the field of intelligent transportation systems (ITS) has achieved remarkable success, which is mainly due to the large amount of available annotation data. However, obtaining these annotated data has to afford expensive costs in reality. Therefore, a more realistic strategy is to leverage semi-supervised learning (SSL) with a small amount of labeled data and a large amount of unlabeled data. Typically, semantic consistency regularization and the two-stage learning methods of decoupling feature extraction and classification have been proven effective. Nevertheless, representation learning only limited to semantic consistency regularization may not guarantee the separation or discriminability of representations of samples with different semantics; due to the inherent limitations of the two-stage learning methods, the extracted features may not match the specific downstream tasks. In order to deal with the above drawbacks, this paper proposes an end-to-end deep semi-supervised learning double contrast of semantic and feature, which extracts effective tasks specific discriminative features by contrasting the semantics/features of positive and negative augmented samples pairs. Moreover, we leverage information theory to explain the rationality of double contrast of semantics and features and slack mutual information to contrastive loss in a simpler way. Finally, the effectiveness of our method is verified in benchmark datasets.
\end{abstract}

\section{Introduction}
Fueled by large-scale of annotated data, supervised deep learning has witnessed promising progress in Computer Vision (CV) \cite{esteva2021deep}, Natural Language Processing (NLP) \cite{lauriola2021introduction} and ITS \cite{guerrero2021deep}.

In recent years, a development trend of ITS is to collect a large amount of data from various sources, and its availability may lead to a major challenge, $i.e.$, the transformation from a traditional technology-driven system to a multi-functional data-driven ITS \cite{kaffash2021big}.
However, in reality, these data consist to a large amount of unlabeled and an extremely small amount of labeled data. Therefore, this brings a huge challenge to the use of ITS.
To meet this challenge, researchers have developed a series of semi-supervised learning methods based on deep networks.
According to their implementation manners, these methods can be roughly divided into the following two categories:
1) two-stage learning pipeline \cite{chen2020big}, that is, feature learning and classification learning are carried out independently one after another. The advantage is that the learned features are relatively more general, while the obvious disadvantage is that the learned features may not be suitable for specific downstream tasks.
2) End-to-end one-stage learning pipeline, that is, joint modeling of features and classification to achieve more task-oriented learning. Typically, such as the regularization method to achieve semantic consistency based on manifold hypothesis and smoothing hypothesis \cite{rasmussemi,laine2016temporal,sajjadi2016regularization,tarvainenmean,yang2021survey}, so that similar input samples have similar or semantically consistent class distribution outputs.
Nevertheless, only constraining semantic consistency may not guarantee the inter class discrimination of feature representation.
Therefore, in order to solve such problems, the motivation of this paper is inspired by the prototype contrast learning method \cite{li2020prototypical} (PCL), and proposes an end-to-end deep semi-supervised learning with ideal discriminant representation.
Specifically, this paper first constructs the contrastive learning paradigm of positive and negative sample pairs at the feature and semantic levels respectively, in order to obtain discriminant features and semantic consistency.
Then, with the help of information theory, it is theoretically explained that contrastive loss can be used as a substitute for loss of mutual information.
On this basis, the optimized contrast loss promotes the maximization of mutual information of positive sample pairs, and theoretically ensures the rationality of the method.
Finally, the effectiveness of the proposed model is verified in benchmark datasets.

Our contributions are summarized as follows:
\begin{itemize}
    \item{ A simple and effective end-to-end semantics and features double-contrast semi-supervised learning method is proposed, which makes the representation of data in feature space achieve better separation between heterogeneous samples.}
    \item{Prove the relationship between the lower bound of mutual information and contrast loss, and provide a theoretical guarantee for the effectiveness of double-contrast.}
    \item{Verify the effectiveness of the proposed method on the benchmark datasets.}
\end{itemize}

\section{Related Works}

Deep semi-supervised learning generally adopts semantic consistency learning based on manifold hypothesis and smoothness hypothesis.
Typical examples are Ladder Network \cite{rasmussemi}, $\Pi$ model \cite{sajjadi2016regularization}, Temporal Ensembling \cite{laine2016temporal}, Mean Teacher \cite{tarvainenmean} etc.
This type of method mainly uses small perturbation on the sample to make the model's semantics or class distribution prediction results as consistent as possible \cite{verma2020interpolation}.
However, these methods only consider the semantics consistency of constrained samples, while ignoring the inter-class relationships of features representation.

In this paper, we introduce contrast learning to solve the problem of semantics and features inconsistency.
Contrastive learning, as a successful unsupervised learning paradigm, aims to extract discriminative features that are as invariant as possible through self-supervised comparison methods for downstream tasks.
Its implementation can be divided into the following two ways: 1) Combining negative examples, that is, extracting sample features by constructing and comparing positive/negative sample pairs.
Typically, MoCo \cite{he2020momentum} achieves the contrast of positive and negative examples by maintaining the representation dictionary;
SimCLR \cite{chen2020simple} directly contrasts the data in the mini-batch to achieve the effect of similar samples close.
PCL \cite{li2020prototypical} explored the difference characteristics of example pairs and contrastive learning between pair-wise and clusters-wise.
2) There is no need for negative cases. Typically, BYOL \cite{grillbootstrap} achieves similar learning by minimizing the loss of two view feature representations generated by the same sample;
SwAV \cite{caron2020unsupervised} can predict each other by using different view feature representations and clustering results generated by the same sample to make the model extract the discriminative feature of invariance.
To the best of our knowledge, this is the first proposal to use double-contrast learning of semantics and features to solve the problem of inconsistent semantic features.

\section{The Proposed Method}
In this section, we regard double-contrast as unsupervised loss, which aims to make the of different semantic samples in the feature space discriminative, and simultaneously make the representation of samples with similar semantics have similar class distribution output.
\subsection{Preliminaries}
Suppose $\boldsymbol{X}=\{\boldsymbol{x_1}, ..., \boldsymbol{x_n}\} \in \mathcal{R}^{D \times n}$ is a set of samples, and, $\boldsymbol{X'}$, witch is augmented by rotation or random blur, is the augmentation of the $\boldsymbol{X}$.
$\phi(\cdot;\theta)$ is the representation network, $g(\cdot;\omega)$ is the classification function.
Representation of the origin samples is $\boldsymbol{Z} = \phi(X;\theta) = \{\boldsymbol{z_1}, ..., \boldsymbol{z_n}\} \in \mathcal{R}^{p \times n}$, correspondingly, the augmented samples representations are $\boldsymbol{Z'} = \phi(\boldsymbol{X'};\theta) = \{\boldsymbol{z^{'}_1}, ..., \boldsymbol{z^{'}_n}\} \in \mathcal{R}^{p \times n}$, among witch $z^{'}_i$ is the positive point of the $z_i$ and rest are negative points.
The set $(\boldsymbol{X}, \boldsymbol{X'})$ constructs the whole training set.
\subsection{Double-contrast learning}

Semi-supervised learning is a promising learning paradigm utilizing small amount of labelled data and huge size of unlabelled data.  Its optimization objectives can be expressed as follows:
\begin{equation}
    \mathcal{L}_{semi} = \mathcal{L}_{sup}(\mathbf{x}, \mathbf{y}; \theta) + \mathcal{L}_{unsup}(\mathbf{x}; \theta)
\end{equation}
where $\mathcal{L}_{sup}(\mathbf{x}, \mathbf{y}; \theta)$ represents the loss of the supervised part ($e.g.,$ cross entropy $\mathcal{L}_{CE}$), and $\mathcal{L}_{unsup}(\mathbf{x}; \theta)$ is the loss of the unsupervised part.

Here, we further treat the double contrast loss as an unsupervised loss. For this reason, we discuss from the perspective that different samples are discriminative and samples with similar characteristics have similar semantic output (class distribution). We first briefly describe these three contrasting losses as follows:

(1)  Feature contrast.
To realize the discriminative representation of different samples in the feature space, the model is transformed into $p\left(\mathbf{z}_{i}^{\prime} \mid \mathbf{z}_{i}\right)>p\left(\mathbf{z}_{\boldsymbol{i}}^{\prime} \mid \mathbf{z}_{i}, j \neq i\right)$.
Then, we maximize the mutual information $MI(\boldsymbol{Z}, \boldsymbol{Z}^{\prime})$ through the following definition:
\begin{equation}
MI\left(\boldsymbol{Z}, \boldsymbol{Z}^{\prime}\right)=\sum_{i=1}^{n} p\left(\mathbf{z}_{\boldsymbol{i}}, \mathbf{z}_{\boldsymbol{i}}^{\prime}\right) \log \frac{p\left(\mathbf{z}_{i}^{\prime} \mid \mathbf{z}_{i}\right)}{p\left(\mathbf{z}_{i}^{\prime}\right)}
\end{equation}.

(2) Semantic contrast.
To maintain the consistency of class distribution output.
For $\boldsymbol{P}=\left[\begin{array}{c}\boldsymbol{q}_{1} \\ \vdots \\ \boldsymbol{q}_{c}\end{array}\right], \quad \boldsymbol{q}_{1} \in \mathcal{R}^{1 \times n}$,  $\boldsymbol{q}_{i}$ is the distribution probability of all samples in the corresponding class of $i$, and the distributions $\boldsymbol{q}'_{i}$ and $\boldsymbol{q}_{i}$ of the corresponding augmented samples are as consistent as possible.
Similarly, we expect the classification function to be transformed into $p\left(\boldsymbol{q}_{i}^{\prime} \mid \boldsymbol{q}_{i}\right)>p\left(\boldsymbol{q}_{j}^{\prime} \mid \boldsymbol{q}_{i}, j \neq i\right)$, and maximize the mutual information $MI(\boldsymbol{q}, \boldsymbol{q}^{\prime})$ through the following definition:

\begin{equation}
MI\left(\boldsymbol{q}, \boldsymbol{q}^{\prime} \right)=\sum_{i=1}^{c} p\left(\boldsymbol{q}_{\boldsymbol{i}}, \boldsymbol{q}_{\boldsymbol{i}}^{\prime}\right) \log \frac{p\left(\boldsymbol{q_{i}}^{\prime} \mid \boldsymbol{q_{i}}\right)}{p\left(\boldsymbol{q_{i}}^{\prime}\right)}
\end{equation}

(3) Double-contrast. To achieve the dual goals of discriminative features and semantic consistency, we jointly define feature and semantic dual mutual information as unsupervised regularization items,
\begin{equation}
\mathcal{L}_{\text {unsup }}(\boldsymbol{x} ; \vartheta)=-M I\left(\boldsymbol{Z}, \boldsymbol{Z}^{\prime}\right)-M I\left(\boldsymbol{q}, \boldsymbol{q}^{\prime}\right)
\end{equation}

Then, we use cross entropy ($\mathcal{L}_{C E}$) as the loss of the supervision part to solve the total loss:
\begin{equation}
\mathcal{L}_{\text {Total }}=\mathcal{L}_{C E}-M I\left(\boldsymbol{Z}, \boldsymbol{Z}^{\prime}\right)-M I\left(\boldsymbol{q}, \boldsymbol{q}^{\prime}\right)
\end{equation}

Contrast loss as a lower bound replacement of mutual information has almost become one of the effective standard methods  \cite{hjelm2018learning}. However, it is very difficult to directly optimize $MI(\boldsymbol{.}, \boldsymbol{.})$. To solve this problem, we relax the mutual information term into contrast loss.

Let $\boldsymbol{R}=\left\{\boldsymbol{r}_{1}, \ldots, \boldsymbol{r}_{n}\right\}$ be a set of samples of random variable $\boldsymbol{r}$, and $R^{\prime}=\left\{\boldsymbol{r}_{1}^{\prime}, \ldots, \boldsymbol{r}_{n}^{\prime}\right\}$ be the corresponding $\boldsymbol{R}$ transform samples. To facilitate the derivation of our further hypotheses, as follows:

\begin{equation}
 f\left(\boldsymbol{r}_{i}, \boldsymbol{r}_{j}^{\prime}\right)=k \cdot \frac{p\left(\boldsymbol{r}_{j}^{\prime} \mid \boldsymbol{r}_{i}\right)}{p\left(\boldsymbol{r}_{j}^{\prime}\right)}
\end{equation}
where $k$ is the constant. we define the contrast loss as:
\begin{equation}
\mathcal{L}_{\text {const }}=-\frac{1}{n} \sum_{i=1}^{n} \log \frac{f\left(\boldsymbol{r}_{i}, \boldsymbol{r}_{i}^{\prime}\right)}{\sum_{j} f\left(\boldsymbol{r}_{i}, \boldsymbol{r}_{j}^{\prime}\right)}
\end{equation}

To illustrate the rationality of the relaxation from Eq. (2) and (3) to $\mathcal{L}_{\text {const}}$, we define the following propositions and prove them.

\textbf{Proposition 1}.
\begin{equation}
\mathcal{L}_{\text {const }} \geq-M I\left(\boldsymbol{R}, \boldsymbol{R}^{\prime}\right)+C
\end{equation}

The calculation result of Eq.8 above is consistent with that of \cite{zhong2020deep}, where $-\mathcal{L}_{\text {const}}$ is the lower bound of mutual information (with a difference of at most one additive constant $C$). The proof is as follows:

\begin{proof}
    \begin{equation}
        \begin{aligned}
            -\frac{1}{n} \sum_{i=1}^{n} &\log \frac{f\left(\boldsymbol{r}_{i}, \boldsymbol{r}_{i}^{\prime}\right)}{\sum_{j} f\left(\boldsymbol{r}_{i}, \boldsymbol{r}_{j}^{\prime}\right)}\\
            &=\frac{1}{n} \sum_{i=1}^{n} \log \left(1+\frac{\sum_{j \neq i} f\left(\boldsymbol{r}_{i}, \boldsymbol{r}_{j}^{\prime}\right)}{f\left(\boldsymbol{r}_{i}, \boldsymbol{r}_{i}^{\prime}\right)}\right)\\
            &\approx \frac{1}{n} \sum_{i=1}^{n} \log\left(1+\frac{(n-1) \mathbb{E}_{r^{'}}\left[f\left(\boldsymbol{r}_{i},\boldsymbol{r}^{\prime}\right)\right]}{f\left(\boldsymbol{r}_{i},\boldsymbol{r}_{i}^{\prime}\right)}\right)\\
            &\geq \frac{1}{n} \sum_{i=1}^{n} \log \left(\frac{(n-1) k}{f\left(\boldsymbol{r}_{i}, \boldsymbol{r}_{i}^{\prime}\right)}\right)\\
            &=-MI\left(\boldsymbol{R}, \boldsymbol{R}^{\prime}\right)+C
        \end{aligned}
    \end{equation}
\end{proof}

From this proposition, we prove that the minimization of contrast loss is equal to the maximization of mutual information, and there is discrimination between positive and negative samples in contrast loss.

Next, we use Proposition 1 to directly derive the slackness of the contrast loss corresponding to the feature and semantic mutual information such as:

(1) Feature contrast loss relaxation.
\begin{equation}
\mathcal{L}_{z}=-\frac{1}{n} \sum_{i=1}^{n} \log \frac{e^{z_{i}^{T} z_{i}^{\prime} / \tau_{f}}}{\sum_{j} e^{z_{i}^{T} z_{j}^{\prime} / \tau_{f}}}
\end{equation}
where $\tau_{f}$  represents the temperature of feature contrast.

(2) Semantic contrast loss relaxation.
\begin{equation}
\mathcal{L}_{q}=-\frac{1}{c} \sum_{i=1}^{c} \log \frac{e^{q_{i} q_{i}^{\prime^{T}} / \tau_{s}}}{\sum_{j} e^{q_{i} q_{j}^{\prime T} / \tau_{s}}}
\end{equation}
where $\tau_{s}$ represents the temperature of semantic contrast.

Finally, we combine the relaxed features and semantic contrast loss to learn total loss of the entire network, as follows:
\begin{equation}
\tilde{\mathcal{L}}_{\text {Total }}=\mathcal{L}_{C E}+\mathcal{L}_{z}+\mathcal{L}_{q}
\end{equation}

\section{Experiments}
In this section, we report the results of tow public datasets to validate the effectiveness of the proposed method.

\noindent\textbf{Datasets}.
We conduct experiments on two widely used datasets of different scenarios, including:

{CIFAR-10} \cite{krizhevsky2009learning}, which contains a natural color image data set of 10 categories of airplanes, cars, birds, cats, deer, dogs, frogs, horses, boats, and trucks. The training set has 5000 samples for each category. A total of 50,000 samples, 1,000 samples for each category in the test set, a total of 10,000 samples, the sample size is 32×32;

SVHN \cite{van2008visualizing} is a digital image data set of 0-9 house numbers, including 10 classes. The training set has 73258 samples and the test set has 26,032 samples. The sample size is 32×32.

\noindent\textbf{Compared Methods}. We compare our method with the following methods, as follows:

We use standard supervised only as the baseline, and compare it with the most popular semantic consistency methods today, including: Ladder Network, $\Pi$ model, Temporal Ensembling, Mean Teacher and VAT.

\noindent\textbf{Implementation}.
In the experiment, we used the backbone network in \cite{tarvainen2017mean}.
For optimization, we first adopt a stochastic gradient descent (SGD) optimizer.
The model is trained for a total of 1000 epochs, and a de-weight of 0.1 is performed at the 500th and 750th epochs.
Secondly, the hyper-parameters are set as follows: the initial learning rate is set to 0.1, the learning rate decay is set to 0.0001, and the SGD momentum is set to 0.9.
During the retraining process, the batch size is set to 512, and the semantic contrast temperature parameter $\tau_{s}=0.9$, and the feature contrast temperature parameter $\tau_{f}=0.5$. Finally, the entire model is implemented by Pytorch.

\begin{figure} [h] \centering
\includegraphics[width=0.49\columnwidth]{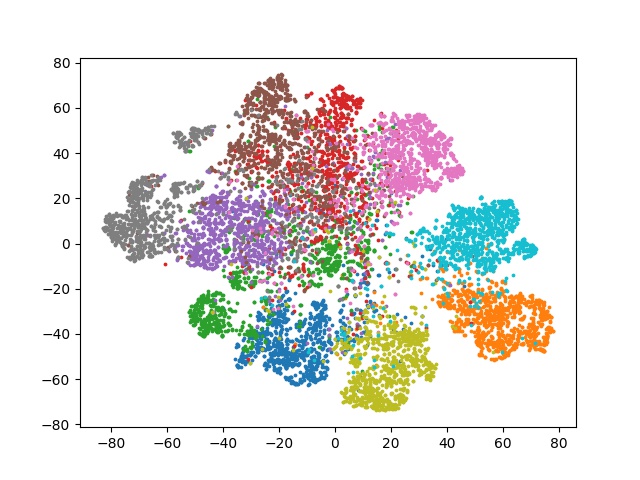}
\includegraphics[width=0.49\columnwidth]{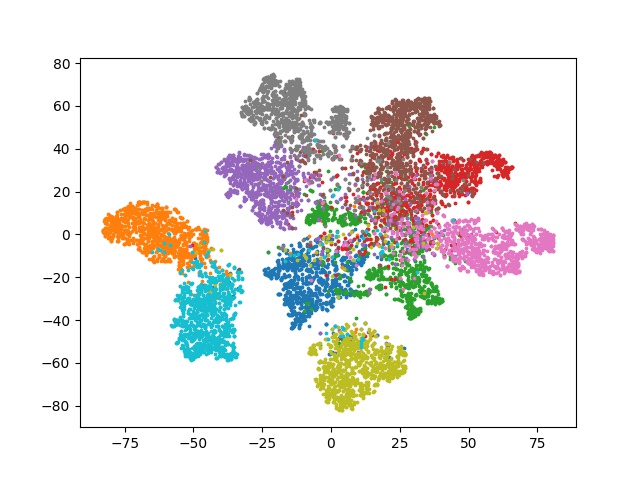}
\caption{Visualize the features of the test set under the constraints of without features comparison and features/semantics double comparison.}
\label{fig}
\end{figure}

\subsection{Comparison Results}

We have conducted experiments on the CIFAR-10 and SVHN datasets respectively, and the results are analyzed as follows:

First of all, our proposed method is better than a single semantic consistency method in performance. The results in Table 1 and Table 2 show that our method has the best performance; Secondly, in the relatively more complex CIFAR-10, our method performs better than other methods, which further shows that the use of double contrast is better suited for relatively complex scenarios. Finally, we use T-SNE \cite{van2008visualizing} to visualize the features extracted by only semantically consistent and double contrast models. As shown in Figure 1, the figure on the right shows that the features extracted by the model under the double contrast loss have better discriminative properties.

\begin{table}[ht]
	\centering
	\caption{Experiment results on CIFAR-10 (classification accuracy)}
	\scriptsize
	\label{exp:1}
	\renewcommand\arraystretch{1.2}
	\setlength{\tabcolsep}{1.2mm}{
		
		\begin{tabular}{c|c|c|c|cc}
			\hline\hline
			\multicolumn{1}{c}{Models}&
			\multicolumn{1}{|c}{{1000 labels}}&
			\multicolumn{1}{|c}{{4000 labels}}&
			\multicolumn{1}{|c}{{50000 labels}}&
			\multicolumn{1}{|c}{\textbf{Avg}}\\\hline
			Supervised Only &$53.6$&$79.3$ &$94.2$& $75.7$&\\
			$\Pi$ mode & $72.6$ & $86.8$ & $93.9$ & $84.4$ & \\
			Mean Teacher &$78.4$ & $87.7$ & $94.0$&$86.7$ &\\
			VAT &$-$&$85.1$&$94.2$&$-$ &\\
			(Ours)&$78.3$ & $\mathbf{89.2}$ & $\mathbf{94.4}$ & $\mathbf{87.3}$ &\\\hline\hline
	    \end{tabular}}
\end{table}

\begin{table}[ht]
	\centering
	\caption{Experiment results on SVHN (classification accuracy)}
	\scriptsize
	\label{exp:1}
	\renewcommand\arraystretch{1.2}
	\setlength{\tabcolsep}{1.2mm}{
		
		\begin{tabular}{c|c|c|c|cc}
			\hline\hline
			\multicolumn{1}{c}{Models}&
			\multicolumn{1}{|c}{{250 labels}}&
			\multicolumn{1}{|c}{{500 labels}}&
			\multicolumn{1}{|c}{{1000 labels}}&
			\multicolumn{1}{|c}{\textbf{Avg}}\\\hline
			Supervised Only &$72.2$&$83.1$ &$87.7$& $81.0$&\\
			$\Pi$ mode &$90.3$ & $93.2$ &$95.1$& $92.8$ & \\
			Mean Teacher &$95.6$&$95.8$&$96.1$&$95.8$ &\\
			VAT &$-$&$-$&$94.6$&$-$ &\\
			(Ours)&$\mathbf{95.6}$&$\mathbf{95.9}$&$96.0$ &$\mathbf{95.8}$&\\\hline\hline
	    \end{tabular}}
\end{table}

\begin{figure}
	\centering
	\includegraphics[width=0.7\linewidth]{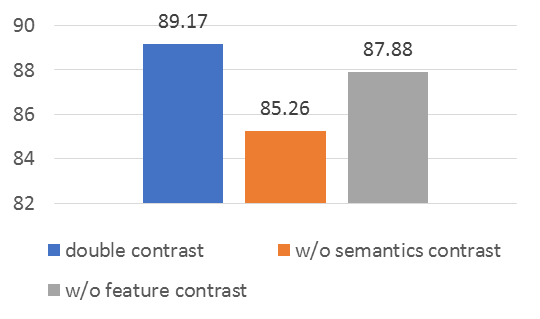}
	\label{fig:select}
	\caption{Performance comparison when eliminating different regularization item.}
\end{figure}

\subsection{Ablation Study}
In this section, we use ablation experiments to further show the importance of the dual contrast of features and semantics to the model.
As shown in Figure 2, we removed semantic contrast (w/o semantic contrast) and feature contrast (w/o feature contrast) when training CIFAR-10 to observe the performance of the model on the test set.
In addition, in conjunction with the results in Table 2, we observe that the dual comparison of semantics and features has better performance.


\subsection{Model convergence analysis}
From the left image of Figure 3 (total loss), right image (feature contrast loss) and the left image of Figure 4 (semantic contrast loss), it can be seen that the loss curve tends to be stable after about 500 epoch iteration training; Figure 4 right image (Cross entropy loss), when the sample label size is 1000, the model is trained for about 700 epoch iterations, and the loss curve tends to stabilize and converge. When the sample size is 4000 labeled data and fully labeled data, the cross-entropy loss curve tends to converge after the model is trained for about 600 epoch iterations. The above experimental results show that the feature semantic double contrast method has better convergence under different labeled data.
\begin{figure} [ht]
\includegraphics[width=0.45\columnwidth]{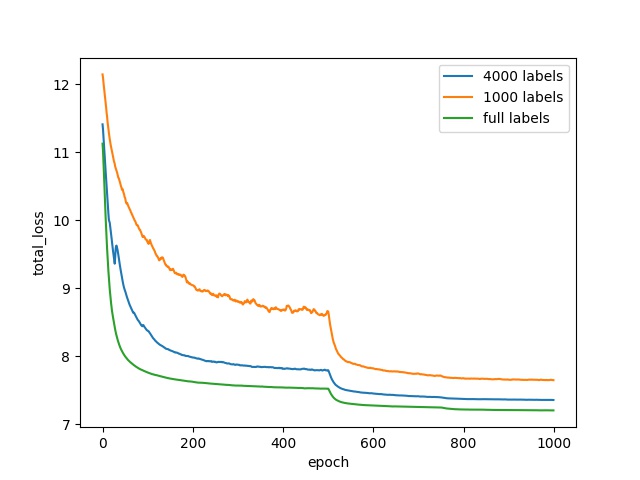}
\includegraphics[width=0.45\columnwidth]{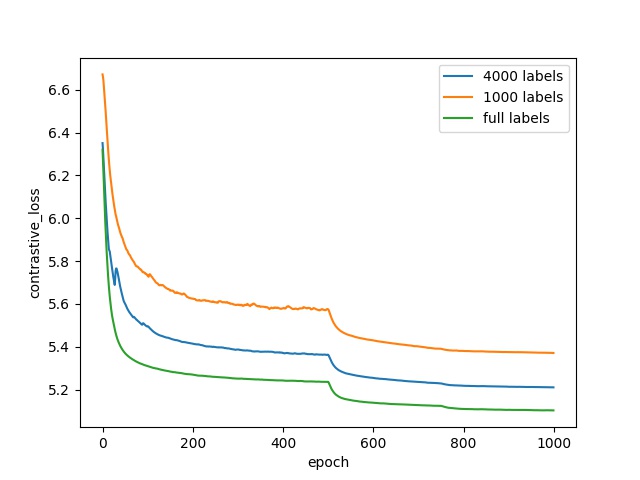}
\caption{The curves of the total loss (left) and feature contrastive loss (right) with different labeled data on the CIFAR-10.}
\label{fig}
\end{figure}

\begin{figure} [ht]
\includegraphics[width=0.45\columnwidth]{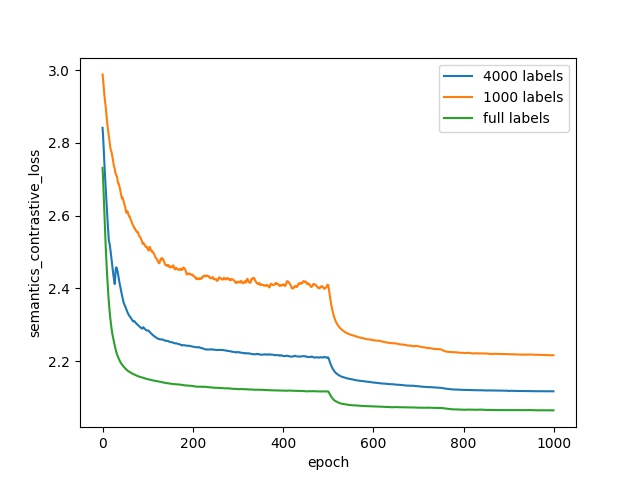}
\includegraphics[width=0.45\columnwidth]{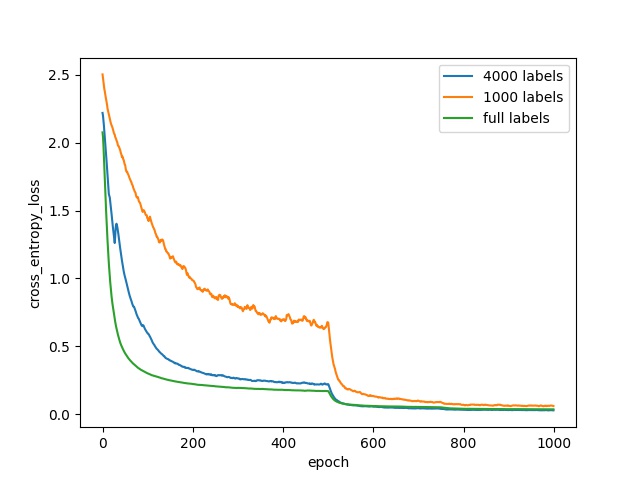}
\caption{The curves of the semantics contrastive loss (left) and cross entropy loss (right) with different labeled data on the CIFAR-10.}
\label{fig}
\end{figure}

\subsection{Conclusion}
This paper first proposes an end-to-end deep semi-supervised model with double contrast of semantics and features. Then, it discusses the use of semantic and feature double contrast model to extract discriminative features suitable for the task.Third, it theoretically proves the hypothesis that contrast loss is used as a mutual information agency loss. Finally, the effectiveness of the double contrast method of semantics and features is verified on the benchmark datasets.

\section{Acknowledgments}
We would like to thank the anonymous reviewers for their valuable comments. This work was supported by National Natural Science Foundation of China (No. 62076251).

\newpage
\bibliographystyle{aaai}
\bibliography{formatting-instructions-latex-2020.blb}

\begin{thebibliography}{}

\bibitem[\protect\citeauthoryear{Caron \bgroup et al\mbox.\egroup
  }{2020}]{caron2020unsupervised}
Caron, M.; Misra, I.; Mairal, J.; Goyal, P.; Bojanowski, P.; and Joulin, A.
\newblock 2020.
\newblock Unsupervised learning of visual features by contrasting cluster
  assignments.
\newblock {\em arXiv preprint arXiv:2006.09882}.

\bibitem[\protect\citeauthoryear{Chen \bgroup et al\mbox.\egroup
  }{2020a}]{chen2020simple}
Chen, T.; Kornblith, S.; Norouzi, M.; and Hinton, G.
\newblock 2020a.
\newblock A simple framework for contrastive learning of visual
  representations.
\newblock In {\em International conference on machine learning},  1597--1607.
\newblock PMLR.

\bibitem[\protect\citeauthoryear{Chen \bgroup et al\mbox.\egroup
  }{2020b}]{chen2020big}
Chen, T.; Kornblith, S.; Swersky, K.; Norouzi, M.; and Hinton, G.
\newblock 2020b.
\newblock Big self-supervised models are strong semi-supervised learners.
\newblock {\em arXiv preprint arXiv:2006.10029}.

\bibitem[\protect\citeauthoryear{Esteva \bgroup et al\mbox.\egroup
  }{2021}]{esteva2021deep}
Esteva, A.; Chou, K.; Yeung, S.; Naik, N.; Madani, A.; Mottaghi, A.; Liu, Y.;
  Topol, E.; Dean, J.; and Socher, R.
\newblock 2021.
\newblock Deep learning-enabled medical computer vision.
\newblock {\em NPJ digital medicine} 4(1):1--9.

\bibitem[\protect\citeauthoryear{Grill \bgroup et al\mbox.\egroup
  }{2020}]{grillbootstrap}
Grill, J.-B.; Strub, F.; Altch{\'e}, F.; Tallec, C.; Richemond, P.~H.;
  Buchatskaya, E.; Doersch, C.; Pires, B.~A.; Guo, Z.~D.; Azar, M.~G.; et~al.
\newblock 2020.
\newblock Bootstrap your own latent: A new approach to self-supervised
  learning.
\newblock {\em arXiv preprint arXiv:2006.07733}.

\bibitem[\protect\citeauthoryear{Guerrero-Iba{\~n}ez, Contreras-Castillo, and
  Zeadally}{2021}]{guerrero2021deep}
Guerrero-Iba{\~n}ez, J.; Contreras-Castillo, J.; and Zeadally, S.
\newblock 2021.
\newblock Deep learning support for intelligent transportation systems.
\newblock {\em Transactions on Emerging Telecommunications Technologies}
  32(3):e4169.

\bibitem[\protect\citeauthoryear{He \bgroup et al\mbox.\egroup
  }{2020}]{he2020momentum}
He, K.; Fan, H.; Wu, Y.; Xie, S.; and Girshick, R.
\newblock 2020.
\newblock Momentum contrast for unsupervised visual representation learning.
\newblock In {\em Proceedings of the IEEE/CVF Conference on Computer Vision and
  Pattern Recognition},  9729--9738.

\bibitem[\protect\citeauthoryear{Hjelm \bgroup et al\mbox.\egroup
  }{2018}]{hjelm2018learning}
Hjelm, R.~D.; Fedorov, A.; Lavoie-Marchildon, S.; Grewal, K.; Bachman, P.;
  Trischler, A.; and Bengio, Y.
\newblock 2018.
\newblock Learning deep representations by mutual information estimation and
  maximization.
\newblock In {\em International Conference on Learning Representations}.

\bibitem[\protect\citeauthoryear{Kaffash, Nguyen, and
  Zhu}{2021}]{kaffash2021big}
Kaffash, S.; Nguyen, A.~T.; and Zhu, J.
\newblock 2021.
\newblock Big data algorithms and applications in intelligent transportation
  system: A review and bibliometric analysis.
\newblock {\em International Journal of Production Economics} 231:107868.

\bibitem[\protect\citeauthoryear{Krizhevsky, Hinton, and
  others}{2009}]{krizhevsky2009learning}
Krizhevsky, A.; Hinton, G.; et~al.
\newblock 2009.
\newblock Learning multiple layers of features from tiny images.

\bibitem[\protect\citeauthoryear{Laine and Aila}{2016}]{laine2016temporal}
Laine, S., and Aila, T.
\newblock 2016.
\newblock Temporal ensembling for semi-supervised learning.
\newblock {\em arXiv preprint arXiv:1610.02242}.

\bibitem[\protect\citeauthoryear{Lauriola, Lavelli, and
  Aiolli}{2021}]{lauriola2021introduction}
Lauriola, I.; Lavelli, A.; and Aiolli, F.
\newblock 2021.
\newblock An introduction to deep learning in natural language processing:
  Models, techniques, and tools.
\newblock {\em Neurocomputing}.

\bibitem[\protect\citeauthoryear{Li \bgroup et al\mbox.\egroup
  }{2020}]{li2020prototypical}
Li, J.; Zhou, P.; Xiong, C.; and Hoi, S.
\newblock 2020.
\newblock Prototypical contrastive learning of unsupervised representations.
\newblock In {\em International Conference on Learning Representations}.

\bibitem[\protect\citeauthoryear{Rasmus \bgroup et al\mbox.\egroup
  }{2015}]{rasmussemi}
Rasmus, A.; Berglund, M.; Honkala, M.; Valpola, H.; and Raiko, T.
\newblock 2015.
\newblock Semi-supervised learning with ladder networks.
\newblock {\em Advances in Neural Information Processing Systems}
  28:3546--3554.

\bibitem[\protect\citeauthoryear{Sajjadi, Javanmardi, and
  Tasdizen}{2016}]{sajjadi2016regularization}
Sajjadi, M.; Javanmardi, M.; and Tasdizen, T.
\newblock 2016.
\newblock Regularization with stochastic transformations and perturbations for
  deep semi-supervised learning.
\newblock {\em Advances in neural information processing systems}
  29:1163--1171.

\bibitem[\protect\citeauthoryear{Tarvainen and Valpola}{2017a}]{tarvainenmean}
Tarvainen, A., and Valpola, H.
\newblock 2017a.
\newblock Mean teachers are better role models: Weight-averaged consistency
  targets improve semi-supervised deep learning results.
\newblock  1195--1204.

\bibitem[\protect\citeauthoryear{Tarvainen and
  Valpola}{2017b}]{tarvainen2017mean}
Tarvainen, A., and Valpola, H.
\newblock 2017b.
\newblock Mean teachers are better role models: Weight-averaged consistency
  targets improve semi-supervised deep learning results.
\newblock In {\em Proceedings of the 31st International Conference on Neural
  Information Processing Systems},  1195--1204.

\bibitem[\protect\citeauthoryear{Van~der Maaten and
  Hinton}{2008}]{van2008visualizing}
Van~der Maaten, L., and Hinton, G.
\newblock 2008.
\newblock Visualizing data using t-sne.
\newblock {\em Journal of machine learning research} 9(11).

\bibitem[\protect\citeauthoryear{Verma \bgroup et al\mbox.\egroup
  }{2020}]{verma2020interpolation}
Verma, V.; Kawaguchi, K.; Lamb, A.; Kannala, J.; Bengio, Y.; and Lopez-Paz, D.
\newblock 2020.
\newblock Interpolation consistency training for semi-supervised learning.
\newblock {\em stat} 1050:29.

\bibitem[\protect\citeauthoryear{Yang \bgroup et al\mbox.\egroup
  }{2021}]{yang2021survey}
Yang, X.; Song, Z.; King, I.; and Xu, Z.
\newblock 2021.
\newblock A survey on deep semi-supervised learning.
\newblock {\em arXiv preprint arXiv:2103.00550}.

\bibitem[\protect\citeauthoryear{Zhong \bgroup et al\mbox.\egroup
  }{2020}]{zhong2020deep}
Zhong, H.; Chen, C.; Jin, Z.; and Hua, X.-S.
\newblock 2020.
\newblock Deep robust clustering by contrastive learning.
\newblock {\em arXiv e-prints}  arXiv--2008.

\end{thebibliography}

\end{document}